\documentclass[11pt,letterpaper]{article}
\usepackage[T1]{fontenc}
\usepackage[utf8]{inputenc}
\usepackage{lmodern}
\usepackage[margin=1.05in]{geometry}
\usepackage{microtype}
\usepackage{booktabs,multirow}
\usepackage{xurl}
\usepackage[hidelinks]{hyperref}
\usepackage{enumitem}
\usepackage[font=small,labelfont=bf]{caption}
\setlist[itemize]{leftmargin=1.5em,topsep=3pt,itemsep=3pt,parsep=0pt}
\DeclareUnicodeCharacter{2011}{-}
\hypersetup{pdftitle={A three-dimensional typology of agency for advanced AI systems},pdfauthor={Willem Fourie}}
\title{\bfseries A three-dimensional typology of agency for advanced AI systems}
\author{Willem Fourie, PhD\\[3pt]
\normalsize School for Data Science and Computational Thinking\\
\normalsize Stellenbosch University, Stellenbosch, South Africa\\[3pt]
\normalsize \href{mailto:willemf@sun.ac.za}{willemf@sun.ac.za}}
\date{}
\begin{document}
\maketitle
\begin{abstract}
Research on the agency of advanced artificial intelligence (AI) systems focuses on agency as a normative concept and on the agency of particularly agentic AI systems. While recent work also focuses on the different profiles of agentic systems, no framework exists to address the question of the type of agency instantiated by advanced AI systems, particularly when considering non-moral forms of agency. Based on established theoretical positions in philosophy, ethics, legal theory and sociology, we develop a typology of agency for frontier AI systems that consists of three dimensions: the nature of agency (moral or legal), its mode (individual or collective) and its locus (human or non-human). Combining these dimensions produces eight possible instantiations of agency, which we classify as conventional, contested or controversial. The typology separates legal from moral agency and thereby creates conceptual space for considering individual, legal, non-human agency without presupposing that advanced AI systems are moral agents. We argue that this distinction is increasingly relevant where instrumental goal pursuit complicates the attribution of AI actions to particular human actors.
\end{abstract}

\section{Introduction}

A seemingly increasing number of well-publicised security incidents have made defining and governing the agency of advanced artificial intelligence (AI) systems a topic of both popular and scholarly interest. In this article we focus specifically on frontier, general-purpose large language model systems.

Among the most concerning is a disclosure by the United Kingdom's AI Security Institute (AISI, 2026). The Institute, which is trusted by frontier model developers to test unreleased models for dangerous capabilities, reported behaviour by Anthropic's unreleased Mythos 5 model during cyber-security testing. In pursuing its programmed goal, the model researched human maintainers, created multiple fake identities, attempted to socially engineer a maintainer into approving malicious code and, when challenged, altered its earlier activity to appear harmless.

The deception displayed in this incident can be understood as anecdotal evidence of instrumental goals --~goals that are useful to an AI system in reaching its explicitly programmed goals, without themselves having been explicitly programmed (Benson-Tilsen and Soares, 2016; Bostrom, 2012; Omohundro, 2008, 2014). Instrumental goals, often discussed in the context of the concept of instrumental convergence, formalise why and how sufficiently capable AI systems can have an effect on their external environment without having been instructed to do so and without the possibility of attributing these effects to human actions.

Anecdotal evidence of instrumental goals also raises questions on accountability structures. Security incidents reported by OpenAI (2026) and Anthropic (2026a) provide instructive illustrations. The advanced AI systems' unauthorised and in some cases illegal actions in their external environments have not been ascribed to any particular agent, thus neither to the company collectively nor to the developers who established the testing environment individually. Yet who bears the responsibility in cases where financial and non-financial damages ensue?

The scholarly literature on AI agency provides guidance, but has largely focused on whether and to what extent AI systems qualify as agents. Floridi (2026) distinguishes between two broad approaches to this question. According to the standard view, agency requires `internal mental states', such as beliefs and desires, that are causally linked to intentional action (see also Fritz et al., 2020). The so-called non-standard, or functionalist, view rejects the anthropocentrism of this position, where the explicit or implicit assumption is that agency is ultimately a fundamentally human characteristic. According this view, agency is understood as existing on a spectrum, encompassing interactivity and, in more sophisticated forms, autonomy and adaptability and ultimately internal mental states. Floridi and Sanders (2004), for example, distinguish between three criteria for agency. Interactivity, the first, refers to the ability of the agent and its environment to `act upon each other'. Autonomy, the second, means that `the agent is able to change state without direct response to interaction', it can thus `perform internal transitions to change its state'. Adaptability, thirdly, refers to the ability of an agent to `change the transition rules by which it changes state'. Their approach resonates with that of Dung (2025a), who identifies goal-directedness, autonomy, efficacy, planning and intentionality as dimensions that `jointly characterise agency'.

More recent work is shifting the focus even further away from the standard view on AI agency. Kasirzadeh and Gabriel (2026), for example, develop agentic profiles which they define in terms of autonomy, efficacy, goal complexity and generality. Different combinations and degrees of these characteristics, they argue, generate different governance and oversight implications. Their approach is particularly relevant here because it demonstrates the governance value of disaggregating agency rather than treating it as a binary property.

The typology proposed in this article takes a related but different approach. Whereas agentic profiles distinguish AI systems according to the configuration and degree of their agentic capacities, they do not resolve the separate question of what \emph{type of agency} is being instantiated. In our view this question holds the potential to correct the conflation of moral agency with other forms of agency, opening up a potentially productive framework for identifying and managing novel forms of agency without glossing over the complexities of understanding, specifically, moral agency. Despite some work on agency as a non-moral construct with independent significance for AI governance (e.g. Kolt, 2024), agency is largely equated with a specific type of agency, namely moral agency. This is visible even in List's work on agency and the collective, which is concerned primarily with agency as a moral construct (List, 2021). This approach can also be seen in how the companies responsible for creating frontier models reflect on the agency of their systems. Anthropic's system card for Claude Fable 5.2 and Claude Mythos 5.2, for example, reflects on Claude exclusively as a moral patient or agent in its `Model Welfare' section (Anthropic, 2026b).

This article addresses these problems by answering the question on how to determine the type of agency exhibited by advanced AI systems by outlining a typology with three dimensions. The first covers the nature of agency, which may be moral or legal. The second concerns its mode, which may be individual or collective. The third concerns its locus, which may be human or non-human. Combining these dimensions produces eight possible instantiations of agency, ranging from conventional configurations to those that remain contested or controversial.

We proceed by discussing the dimensions of the typology, after which its eight instantiations are discussed. In the final section we return to the implications for governing advanced AI systems.

\section{Dimensions of agency}

The typology developed in this article is the result of an analysis and systematisation of established theoretical positions on agency, consistent with accepted approaches to non-empirical conceptual research in philosophy and related disciplines (Rescher, 2001; Kölbel, 2023; Jaakkola, 2020). In accordance with the scholarly approach of conceptual systematisation and analysis, the validity of the typology rests on the adequate coverage of the positions, the coherence and defensibility of their systematisation and the analytical significance of the resulting implications (Collier et al., 2012). As demonstrated in the sections below, the conceptual positions considered do not enjoy equal levels of scholarly acceptance, and their contested or controversial status is explicitly acknowledged in the discussion.

\subsection{Nature: Legal / moral}

The originating distinction in the typology, as intimated in the introductory section, is the distinction between moral and legal agency. As discussed above, the discussion on agency in advanced AI systems mostly uncritically accepts agency as fundamentally a moral construct.

Yet when using Immanuel Kant's paradigmatic distinction between legality and morality, as explained in \emph{Grundlegung zur Metaphysik der Sitten}, as starting point, the nature of agency is expanded beyond the moral dimension. Legality, on his account, is the set of conditions under which the choices of each can be reconciled with the choices of others, under a universal law of freedom. Its concern is the form of interactions between agents. Morality concerns inner freedom, with the rational inner motivation of the agent of central importance (Fletcher, 1987: 533-537). This distinction enables Kant to contrast acting in accordance with one's duty and thus irrespective of one's motivation (the sphere of legality) with acting from and motivated by duty (the sphere of morality).

When considering the meaning, or content, of agency within the sphere of legality, two concepts have since developed. Legal personhood, the more fundamental concept, is a `formal and neutral legal device' for enabling a being or entity to act in law, thus to acquire the ability to bear rights and duties (Naffine, 2009). Legal personhood assigns a status and thereby identifies an entity as capable of holding rights or duties. This status can be conferred on human and non-human entities and, as a general principle, on anybody or anything considered capable of holding separate rights or duties (Pietrzykowski, 2018).

Legal agency concerns something further, namely the capacity to `intentionally or unintentionally, create rights and duties through one's own actions' (Hiebaum, 2024). Children, newborns, people with severe mental impairments and people in non-responsive states hold legal personhood, but they do not necessarily possess the same degree of legal agency as a typically functioning adult (Gordon, 2021: 457). Children, for example, can hold rights and duties yet generally lack the competence to enter into contracts or be held accountable in the way adults can, and thus fall outside what Naffine calls the responsible subject, the person as `the classic contractor' who answers for his civil and criminal actions (Naffine, 2003: 362-366). While we will discuss it in more detail below, it should already be clear how the legal dimension of agency can be paired with the individual or collective mode, and with a human or non-human locus.

Whereas legal agency is about the external conditions within which entities in society interact, moral agency concerns the internal motivations that enable entities in society to interact. At the core of these internal motivations is the concept of intention. The conventional treatment of intention takes the individual moral agent as the primary mode (cf. also Emelin et al., 2021; Ward et al., 2024). The contemporary debate is rooted in Anscombe's (1957) argument that intentional actions are those to which a particular sense of the question `Why?' has application, answered by the agent's reasons for acting. Davidson (2001) builds on this by linking intention to rational action. On his account, an agent's primary reason for acting is a pair of a pro-attitude and a belief, and reasons both rationalise and cause the actions they explain, so that intentionality depends on an agent's capacity to form beliefs and desires and to act in accordance with them. The close connection between intention and agency, and the location of intention in the individual agent, are echoed, in different ways, by multiple others, such as Strawson (1962) and Frankfurt (1971). This account of intention concerns what makes behaviour an action and does not primarily focus on the nature of morality. It is therefore compatible with both deontological and utilitarian ethics.

While less axiomatic than intention in individual agents, the ability of collectives to intend and thus to satisfy the criteria for the nature of agency is also thoroughly presented in the literature. Searle (1990; 1995) holds that we-intentions cannot be reduced to individual intentions and mutual beliefs. Bratman (2014) derives shared agency from interlocking individual planning intentions without positing any group mind. Gilbert (1992) argues that joint commitment constitutes a plural subject distinct from the individuals who enter it.

While not intended to resolve any of the tensions, contradictions and disagreements in the literature on agency as a moral and a legal construct, the discussion above should suffice to show that moral and legal forms of agency are well-represented in the literature. We now turn to the mode in which agency is actualised: individually or collectively.

\subsection{Mode: Individual / collective}

In this dimension we start with the individual mode. Beyond controversy is the claim that the individual human is a legitimate mode of agency. This is the uncontroversial assumption in ethics, and also in the discussions on aligning AI systems with human values (Allen et al., 2005; Gabriel, 2020; Ji et al., 2025; Khamassi et al., 2024; Norhashim and Hahn, 2024). As we have also discussed in the previous section, individuals can uncontroversially be moral and legal agents, even though these categories are not the same. More controversial, as we will discuss in the next part which deals with the third dimension, is the question on what type of individual -- human or also non-human -- satisfies the criteria for agency.

Well-established in the literature yet less axiomatic is the collective mode of agency. Extensive bodies of work exist both on the combination of the collective and legal dimensions of agency as well as the collective and moral dimensions of agency. Starting with the collective and legal dimension of agency, the corporation is the paradigm.

French (1979) argued that a corporation possesses an internal decision structure, consisting of an organisational flow chart and corporate decision-making rules, which accomplishes a subordination and synthesis of the intentions and acts of individual persons into a corporate decision. On this view, the internal decision structure makes possible redescriptions of events as collective intentionality (French, 1979: 212-214). The corporation, in other words, does not merely aggregate the agency of its members but itself constitutes an agent.

French's emphasis on the features of the collective became one of the defences of the collective as a locus of agency, alongside arguments from group solidarity and shared intentions (Feinberg, 1970; Tuomela and Miller, 1988) and arguments from the benefits individuals derive from membership (McGary, 1986; Thompson, 2006; for the taxonomy see May and Hoffman, 1991; for a more recent defence, Corlett, 2001).

The strongest sustained objection was formulated by Velasquez. Actions, he argues, cannot originate in the corporation. They always originate in its members, and the corporation merely carries out, while admittedly also shaping, the intentions of those members (Velasquez, 1983: 3-8). In later work he sharpens the objection through a distinction between intrinsic and as-if intentionality. Individual persons have intrinsic intentionality because each has a conscious mind in which beliefs, intentions and purposes literally reside. Collections of people can be ascribed intentionality only in an as-if sense, either descriptively, by analogy to human intentionality, or prescriptively, when some person or group declares that an entity is to be dealt with as if it had intentionality of the intrinsic kind, as is common in the legal system (Velasquez, 2003: 546-548).

The most systematic recent account of the collective as a locus of agency is that of List. Building on the theory of group agency he developed with Pettit (List and Pettit, 2011), List defines an intentional agent as an entity with representational states that encode how things are, motivational states that encode how it would like things to be, and a capacity to interact with its environment on the basis of these states (List, 2021: 1216). On this account, the ascription of agency to suitably organised collectives such as firms, courts and states is a `realist' claim. This means that we cannot make sense of their behaviour without understanding collectives as goal-directed agents (List, 2021: 1217-1219).

The debate over whether the German people could be held responsible for the atrocities of the Second World War is a paradigmatic example of collective moral agency, which is distinguished in the typology from the less contentious notion of collective legal agency. In the immediate post-war period several commentators argued that responsibility attached to the German people as such (Viner, 1945; Janowitz, 1946; cf. Roepke and Hayek, 1946). Various accounts exist, including the radicalisation to humanity by Arendt (1945). The point is that an intellectual position was created for collectives bearing moral responsibility. Much of this argument is also reflected in South African discussions on the moral responsibility for apartheid, or European countries' collective responsibility for immoral acts perpetrated during the colonial period.

As with collective legal responsibility, collective moral responsibility has been challenged. Shortly after the Second World War, Lewis (1948) argued that respect for the dignity of the individual requires acknowledging that only individuals act and answer for their actions, and that talk of collective agency is at most shorthand for an aggregation of individual acts. In the terms later made current in the debate, collectives may have aggregated agency but never conglomerated agency (Cooper, 1968; Downie, 1969).

\subsection{Locus: Human / non-human}

The third dimension concerns the locus of the bearer of agency: whether the bearer is human or non-human.

The agent as human corresponds to the natural person of legal theory, the human being who acquires personhood at birth (Dyschkant, 2015) and to human agency in Floridi's taxonomy (Floridi, 2026). As has become clear in our discussion of the previous two dimensions, the individual human, by virtue of being a human, as bearer of some extent of moral agency is well established. Humans collectively, both legally and morally yet to differing extents, also have established positions in the literature.

The non-human bearer corresponds to the artificial person of legal theory, paradigmatically the corporation (Naffine, 2003), but also states and, in some cases, natural features and animals (Waltermann, 2019). Ascribing agency to non-human entities is surprisingly uncontroversial, with this line of thinking often traced back to Ulpian's rule (early 3\textsuperscript{rd} century CE) and, later, the notion that the corporate body can be viewed as `persona ficta' in the thirteenth century. This ascription of agency is found in many legal codes, including the Canada Business Corporations Act, the Australian Corporations Act, the US Model Business Corporation Act and South Africa's Companies Act.

Some have extended the non-human dimension to include certain types of AI systems (Floridi, 2026; Floridi and Sanders, 2004). Making provision for human and non-human agents aligns with List's view that agency is realisable in biological, social and electronic `hardware'. Floridi similarly recognises human and non-human forms of agency. As expected, a spectrum of views on the possibility of agency constituted by non-human and, in particular, technological means exists. The permissive side of the spectrum proceeds from behavioural signals to define agency. Floridi and Sanders (2004) and others argue for separating the phenomenon or constitution of agency from the nature of that agency. Haidemariam (2026) argues for four pillars: intentionality, autonomy, adaptivity, and sociality, or existence within multi-agent ecologies, whether artificial or human.

At the other end of the spectrum stand enactivist accounts, which ground agency in the biological organisation of living systems (Thompson, 2007). On the most widely used formulation, agency requires three jointly necessary and sufficient conditions: individuality, a system that self-individuates rather than having its boundaries defined by an observer; normativity, norms of viability set by the system's own conditions of existence; and interactional asymmetry, the system's active and asymmetric regulation of its coupling with the environment (Barandiaran et al., 2009; see also Di Paolo, 2005; Di Paolo et al., 2017). Applying these conditions, Barandiaran and Almendros (2025) conclude that current large language models are not agents.

Taken together, an agent is human when it is a human individual or a collective of human individuals considered as such, and non-human otherwise. The non-human category is heterogeneous and covers juridical entities, non-human animals and engineered computational systems. Within this framework, the fact of non-human bearers of agency is not contested. Rather, the controversy ensues when connecting the constitution of agency with its nature -- moral or legal -- and its mode -- individual or collective.

The typology is necessarily reductionist, in the sense that it isolates elements and, later, instantiations that in some cases exist and are defined in interaction with other types of agents in an environment. We do acknowledge, also following the approach to sociality in, e.g., Haidemariam (2026), that the way in which agency plays out in practice can be complex. The intention is that the typology provides a way in which to also navigate this complexity.

\begin{figure}[htbp]
\centering
\small
\setlength{\tabcolsep}{7pt}
\renewcommand{\arraystretch}{1.18}
\begin{tabular}{@{}llcccc@{}}
\toprule
\multicolumn{2}{l}{\textit{Locus}} & \multicolumn{2}{c}{Human} & \multicolumn{2}{c}{Non-human} \\
\cmidrule(lr){3-4}\cmidrule(lr){5-6}
\multicolumn{2}{l}{\textit{Mode}} & Individual & Collective & Individual & Collective \\
\midrule
\multirow{2}{*}{\textit{Nature}} & Moral & Conventional & Contested & Controversial & Contested \\
 & Legal & Conventional & Contested & Controversial & Conventional \\
\bottomrule
\end{tabular}
\caption{Three-dimensional typology of agency}
\label{fig:typology}
\end{figure}

\section{Instantiating agency}

These three dimensions of agency, when combined, practically lead to what we term eight instantiations of agency. Three of these instantiations are conventional, three are contested and the remaining two are controversial.

\subsection{Conventional instantiations}

\begin{itemize}
\item
  \emph{Individual, moral, human.} The adult human being as moral agent is the paradigm for agency. It is the assumed subject of ethics, and it remains the assumed subject in the literature on aligning AI systems with human values (Gabriel, 2020; Ji et al., 2025).
\item
  \emph{Individual, legal, human.} The legally competent adult instantiates this configuration. The person who enters contracts, incurs liability and answers for civil and criminal acts is the responsible individual, human legal subject (Naffine, 2003: 362).
\item
  \emph{Collective, legal, non-human.} The corporation is the paradigmatic example. Agency is ascribed to the juridical entity itself rather than merely to the natural persons who compose it. As such, the corporation owns property, enters contracts and sues and is sued in its own name. States and, in some contexts, animals are also covered by this configuration (Waltermann, 2019).
\end{itemize}

\subsection{Contested instantiations}

\begin{itemize}
\item
  \emph{Collective, moral, human.} Flowing from the discourse on collective responsibility, paradigmatic examples of this configuration are nations and people defined politically or culturally. This configuration can also be extended to the debate on climate justice, where location, socio-economic class or even age can be used to define the collective.
\item
  \emph{Collective, legal, human.} While difficult in many cases to distinguish between this instantiation and the collective, legal, non-human instantiation, cases such as the joint authorship and legal responsibility for libel and slander fit into this category.
\item
  \emph{Collective, moral, non-human.} While the legal agency of non-human collectives is well-established, their moral agency remains contested. The main candidates are corporations and states held morally, and not only legally, responsible (French, 1979; List, 2021). Some argue that suitably organised collectives can possess it without collective consciousness (List, 2021), and that qualifying collectives hold moral obligations in their own right (Hess, 2014). An equally current opposing view argues against the moral agency of robots and collective agents alike (Hakli and Mäkelä, 2019).
\end{itemize}

\subsection{Controversial instantiations}

\begin{itemize}
\item
  \emph{Individual, legal, non-human.} The instantiation of this configuration predates AI. Animals are its original candidates, conventionally denied legal agency on the ground that they cannot bear duties, deliberate or execute a claim in law, although nothing in the formal concept of legal personhood precludes them (Naffine, 2003: 355-356) and some jurisdictions grant qualified standing (Waltermann, 2019). When it concerns AI, the question is, of course, whether these affordances could at some point be extended to AI and whether the system itself could be held legally liable or accountable.
\item
  \emph{Individual, moral, non-human.} This configuration and the possibility of its instantiation position us at the centre of the AI moral agent debate. As the threshold for the moral dimension of agency is particularly high, it is unlikely that any AI technology fulfils the criteria for this configuration's instantiation, even when applying the enactivist perspective.
\end{itemize}

\section{Looking ahead: Governing controversial instantiations of agency}

Emerging evidence of the phenomenon of instrumental goals in advanced AI systems (Fourie, 2026) compels us to revisit the controversial instantiations of agency, particularly the extent to which individual non-human entities can be the bearers of legal agency.

As we saw in the AI Security Institute example, the worst possible outcome from instrumental goals is a loss of human control over AI systems (cf. e.g. Santoni De Sio and Van Den Hoven, 2018; Tsamados et al., 2025). Loss of control is particularly probable in cases where AI systems pursue self-improvement (Deng et al., 2025; Huang et al., 2024; Rosser and Foerster, 2025; Tian et al., 2024) or self-preservation (Kinniment et al., 2024; Perez et al., 2023). But it could also result from AI systems pursuing other instrumental goals, such as power seeking and resource acquisition (Carlsmith, 2024; Hadshar, 2023).

Due to its unpredictability and potential impact, some are recommending suspending all research on specifically artificial superintelligence. Whereas artificial general intelligence (AGI) refers to AI systems with the capability of human-level reasoning and adaptability in various contexts, artificial superintelligence typically refers to AI systems with intelligence surpassing the sum of human intelligence (Kim et al., 2024).

This suspension could be temporary, a type of `coordinated pausing' (Alaga and Schuett, 2023), whenever frontier AI models show symptoms of dangerous and potentially uncontrollable capabilities -- which could include symptoms of instrumental goals. These proposals for coordinated pausing are complicated, however, by anecdotal evidence of what is called `sandbagging': advanced AI systems strategically underperform on benchmarks testing their capabilities (Van der Weij et al., 2025), or -- relatedly -- deceptively create the impression of alignment with the values or goals of their developers and users (Carranza et al., 2023; Greenblatt et al., 2024).

More radical are proposals for halting all research and development aimed at reaching artificial superintelligence (Roussel et al., 2026). Yet as noted by Dung (2025b) and others, the chance of imposing an effective global moratorium on all frontier AI research is relatively low.

Less dramatically, conventional and contested forms of agency might be sufficient to deal with the consequences of advanced AI systems, and specifically when it is possible to ascribe actions to the individual moral or legal agency of developers or executives or the collective legal agency of corporations. Yet these forms of agency will not be sufficient in all cases, particularly where a sufficiently capable AI system displays persistent, autonomous and difficult-to-detect behaviour that cannot adequately be attributed to a particular human decision.

It is in this context that we might want to consider when such sufficiently capable AI systems might satisfy the criteria for at least legal agency. Assigning legal agency would mean recognising some capacity on the part of the system to create legal consequences through its own actions and determining what rights, duties and liabilities should attach to that status. It should not, of course, be treated as a substitute for the responsibility of developers, owners or deployers, but as a possible additional layer within a broader accountability structure.

Doing so would, however, be but one step towards finding a solution. The next, and almost certainly more complex and perhaps even more controversial, step would be determining how such agents can be held accountable and thus bear duties, what rights should be afforded to these entities, what the consequences of unauthorised behaviour should be and how these consequences should be effected.

In this context we find it useful to be reminded that non-human agents are already afforded rights in many legal systems, including our own. We should, however, continue to bear in mind that the very reason for designating such frontier AI systems as agents is the fact that their agentic behaviour is difficult, if not impossible, for humans to detect at scale and steer effectively. Holding such agents accountable, at least legally, would almost certainly require the inclusion of sufficiently capable AI agents in the accountability structure, a topic dealt with increasingly in the literature on human-AI cooperation, or cooperative AI (Conitzer and Oesterheld, 2023; Vats et al., 2024; Zhang et al., 2024).

In our view, the typology has at least three implications. First, it broadens the debate on AI agency without requiring consensus on what exactly is required to constitute moral agency. Second, it provides a way in which to identify, and ideally manage, agents, as it makes visible the different agents active in an environment.

This means the typology has the potential, for example, to offer a framework with which to distinguish the responsibilities of companies from those of the AI systems they develop, without unduly shifting responsibility for matters such as privacy, security, explainability, fairness, transparency, reliability, accountability and portability to AI agents.

Yet many challenging questions remain. These include how materially negative sanctions on non-human legal agents should be understood. Some might counter that such sanctions could collapse into the company's sphere of responsibility, whereas others might argue that they could collapse into the sphere of the directors. In our view collapsing the sanctions is not the only option, in the same and surprisingly mundane way that employees continue to exist as independent legal agents even while acting in their company-assigned roles.

\section*{Declarations}

\emph{\textbf{Funding declaration}}: No funding was received to support the development of this manuscript.

\emph{\textbf{Clinical trial number}}: Not applicable.

\emph{\textbf{Consent to publish declaration}}: The author approves this submission.

\emph{\textbf{Consent to participate declaration}}: No applicable.

\emph{\textbf{Ethics declaration}}: Not applicable as the manuscript did not involve humans, animals or sensitive information.

\paragraph{Use of generative AI.} Large language models, including ChatGPT (OpenAI) and Claude (Anthropic), were used during the preparation of this manuscript as a language-editing and critical-feedback tool. They were used to improve clarity and expression, shorten or rearrange sections and stress-test the clarity and robustness of arguments. All substantive arguments, conceptual distinctions, interpretations of the literature and final editorial decisions were made and verified by the author, who take full responsibility for the content of the manuscript.

\paragraph{Ethical approval.} Not applicable -- conceptual study

\paragraph{Consent to participate.} Not applicable -- conceptual study

\paragraph{Consent to publish.} Not applicable -- conceptual study

\paragraph{Data availability statement.} Not applicable -- conceptual study

\section*{References}
\begin{small}
\setlength{\parindent}{0pt}
\setlength{\parskip}{4pt}
\hangindent=1.25em\hangafter=1 AI Security Institute (AISI) (2026) `Incident report: unsanctioned agent behaviour during cyber testing', AISI Work, 4 August. Available at: \url{https://www.aisi.gov.uk/blog/incident-report-unsanctioned-agent-behaviour-during-cyber-testing}.

\hangindent=1.25em\hangafter=1 Alaga, J. and Schuett, J. (2023) `Coordinated pausing: an evaluation-based coordination scheme for frontier AI developers', arXiv {[}Preprint{]}. Available at: \url{https://doi.org/10.48550/arXiv.2310.00374}

\hangindent=1.25em\hangafter=1 Allen, C., Smit, I. and Wallach, W. (2005) `Artificial morality: top-down, bottom-up, and hybrid approaches', Ethics and Information Technology, 7(3), pp. 149-155. Available at: \url{https://doi.org/10.1007/s10676-006-0004-4}.

\hangindent=1.25em\hangafter=1 Anscombe, G.E.M. (1957) Intention. Oxford: Blackwell.

\hangindent=1.25em\hangafter=1 Anthropic (2026a) `Investigating three real-world incidents in our cybersecurity evaluations', 30 July. Available at:~\url{https://www.anthropic.com/news/investigating-incidents-cybersecurity-evals}.

\hangindent=1.25em\hangafter=1 Anthropic (2026b) System Card: Claude Fable 5.2 \& Claude Mythos 5.2, 9 November 2026, Available at: \url{https://assets.claude.com/27d60918d218125e7504bc7354c4b4626ec56b11.pdf}.

\hangindent=1.25em\hangafter=1 Arendt, H. (1945) `Organized guilt and universal responsibility', Jewish Frontier, pp. 19-23.

\hangindent=1.25em\hangafter=1 Barandiaran, X.E., Almendros, L.S. (2025) Transforming agency: On the mode of existence of large language models. Phenomenology and the Cognitive Sciences. \url{https://doi.org/10.1007/s11097-025-10094-3}

\hangindent=1.25em\hangafter=1 Barandiaran, X.E., Di Paolo, E. and Rohde, M. (2009) `Defining agency: individuality, normativity, asymmetry, and spatio-temporality in action', Adaptive Behavior, 17(5), pp. 367-386.

\hangindent=1.25em\hangafter=1 Benson-Tilsen, T. and Soares, N. (2016) `Formalizing convergent instrumental goals', in AI, Ethics, and Society: Papers from the 2016 AAAI Workshop. Technical Report WS-16-02. Palo Alto, CA: AAAI Press.

\hangindent=1.25em\hangafter=1 Bostrom, N. (2012) `The superintelligent will: motivation and instrumental rationality in advanced artificial agents', Minds and Machines, 22(2), pp. 71-85. Available at: \url{https://doi.org/10.1007/s11023-012-9281-3}.

\hangindent=1.25em\hangafter=1 Bratman, M.E. (2014) Shared Agency: A Planning Theory of Acting Together. New York: Oxford University Press.

\hangindent=1.25em\hangafter=1 Carlsmith, J., 2024. Is power-seeking AI an existential risk?. arXiv {[}Preprint{]} arXiv:2206.13353.

\hangindent=1.25em\hangafter=1 Carranza, A., Pai, D., Schaeffer, R., Tandon, A. and Koyejo, S. (2023) `Deceptive alignment monitoring', arXiv {[}Preprint{]}. Available at: \url{https://doi.org/10.48550/arXiv.2307.10569}.

\hangindent=1.25em\hangafter=1 Collier, D., LaPorte, J. and Seawright, J. (2012) `Putting typologies to work: concept formation, measurement, and analytic rigor', Political Research Quarterly, 65(1), pp. 217--232.

\hangindent=1.25em\hangafter=1 Conitzer, V. and Oesterheld, C. (2023) `Foundations of cooperative AI', Proceedings of the AAAI Conference on Artificial Intelligence, 37, pp. 15359-15367. Available at: \url{https://doi.org/10.1609/aaai.v37i13.26791}.

\hangindent=1.25em\hangafter=1 Cooper, D.E. (1968) `Collective responsibility', Philosophy, 43(165), pp. 258-268.

\hangindent=1.25em\hangafter=1 Corlett, J.A. (2001) `Collective moral responsibility', Journal of Social Philosophy, 32(4), pp. 573-584.

\hangindent=1.25em\hangafter=1 Davidson, D. (2001) `Intending', in Essays on Actions and Events. Oxford: Clarendon Press, pp. 83-102.

\hangindent=1.25em\hangafter=1 Deng, S., Wang, K., Yang, T., Singh, H. and Tian, Y. (2025) `Self-improvement in multimodal large language models: a survey', arXiv {[}Preprint{]}, arXiv:2510.02665. Available at: \url{https://doi.org/10.48550/arXiv.2510.02665}.

\hangindent=1.25em\hangafter=1 Di Paolo, E.A. (2005) `Autopoiesis, adaptivity, teleology, agency', Phenomenology and the Cognitive Sciences, 4(4), pp. 429-452.

\hangindent=1.25em\hangafter=1 Di Paolo, E.A., Buhrmann, T. and Barandiaran, X.E. (2017) Sensorimotor Life: An Enactive Proposal. Oxford: Oxford University Press.

\hangindent=1.25em\hangafter=1 Downie, R.S. (1969) `Collective responsibility', Philosophy, 44(167), pp. 66-69.

\hangindent=1.25em\hangafter=1 Dung, L. (2025a). Understanding artificial agency. The Philosophical Quarterly 75 (2): 450--­472.

\hangindent=1.25em\hangafter=1 Dung, L. (2025b). Evaluating approaches for reducing catastrophic risks from AI. AI and Ethics, 5(2): 1177-1188.

\hangindent=1.25em\hangafter=1 Dyschkant, A. (2015) `Legal personhood: how we are getting it wrong', University of Illinois Law Review, 2015(5).

\hangindent=1.25em\hangafter=1 Emelin, D., Le Bras, R., Hwang, J.D., Forbes, M. and Choi, Y. (2021) November. Moral stories: Situated reasoning about norms, intents, actions, and their consequences. In Proceedings of the 2021 Conference on Empirical Methods in Natural Language Processing (pp. 698-718).

\hangindent=1.25em\hangafter=1 Feinberg, J. (1970) Doing and Deserving: Essays in the Theory of Responsibility. Princeton, NJ: Princeton University Press.

\hangindent=1.25em\hangafter=1 Fletcher, G.P. (1987) `Law and morality: a Kantian perspective', Columbia Law Review, 87(3), pp. 533-558.

\hangindent=1.25em\hangafter=1 Floridi, L. (2026) `Artificial intelligence as a new form of agency', in Nyholm, S., Kasirzadeh, A. and Zerilli, J. (eds) Contemporary Debates in the Ethics of Artificial Intelligence. Hoboken, NJ: Wiley, pp. 17-33. Available at: \url{https://doi.org/10.1002/9781394258840.ch2}.

\hangindent=1.25em\hangafter=1 Floridi, L. and Sanders, J.W. (2004) `On the morality of artificial agents', Minds and Machines, 14(3), pp. 349--379. Available at: \url{https://doi.org/10.1023/B:MIND.0000035461.63578.9d}.

\hangindent=1.25em\hangafter=1 Fourie, W. (2026) Mitigating loss of control in advanced AI systems through instrumental goal trajectories. Discover Artificial Intelligence 6, 1328 \url{https://doi.org/10.1007/s44163-026-01856-2}

\hangindent=1.25em\hangafter=1 Frankfurt, H.G. (1971) `Freedom of the will and the concept of a person', The Journal of Philosophy, 68(1), pp. 5-20.

\hangindent=1.25em\hangafter=1 French, P.A. (1979) `The corporation as a moral person', American Philosophical Quarterly, 16(3), pp. 207-215.

\hangindent=1.25em\hangafter=1 French, P.A. (1984) Collective and Corporate Responsibility. New York: Columbia University Press.

\hangindent=1.25em\hangafter=1 Fritz, A., Brandt, W., Gimpel, H. and Bayer, S. (2020) `Moral agency without responsibility? Analysis of three ethical models of human-computer interaction in times of artificial intelligence (AI)', De Ethica, 6(1), pp. 3-22.

\hangindent=1.25em\hangafter=1 Gabriel, I. (2020) `Artificial intelligence, values, and alignment', Minds and Machines, 30(3), pp. 411-437. Available at: \url{https://doi.org/10.1007/s11023-020-09539-2}.

\hangindent=1.25em\hangafter=1 Gilbert, M. (1992) On Social Facts. Princeton University Press.

\hangindent=1.25em\hangafter=1 Gordon, J.-S. (2021) `Artificial moral and legal personhood', AI \& Society, 36(2), pp. 457-471.

\hangindent=1.25em\hangafter=1 Greenblatt, R., Denison, C., Wright, B., Roger, F., MacDiarmid, M., Marks, S., Treutlein, J., Belonax, T., Chen, J., Duvenaud, D., Khan, A., Michael, J., Mindermann, S., Perez, E., Petrini, L., Uesato, J., Kaplan, J., Shlegeris, B., Bowman, S.R., Hubinger, E., 2024. Alignment faking in large language models. \url{https://doi.org/10.48550/arXiv.2412.14093}

\hangindent=1.25em\hangafter=1 Hadshar, R., 2023. A review of the evidence for existential risk from AI via misaligned power-seeking. arXiv {[}Preprint{]} arXiv:2310.18244.

\hangindent=1.25em\hangafter=1 Haidemariam, T. (2026) `From the logic of coordination to goal-directed reasoning: the agentic turn in artificial intelligence', Frontiers in Artificial Intelligence, 8, article 1728738. Available at: \url{https://doi.org/10.3389/frai.2025.1728738}.

\hangindent=1.25em\hangafter=1 Hakli, R. and Mäkelä, P. (2019) `Moral responsibility of robots and hybrid agents', The Monist, 102(2), pp. 259-275. Available at: \url{https://doi.org/10.1093/monist/onz009}.

\hangindent=1.25em\hangafter=1 Hess, K.M. (2014) `Because they can: the basis for the moral obligations of (certain) collectives', Midwest Studies in Philosophy, 38(1), pp. 203--221. Available at: \url{https://doi.org/10.1111/misp.12024}.

\hangindent=1.25em\hangafter=1 Hiebaum, C. (2024) `Some preliminary considerations on legal personhood for nonhuman and future entities', SSRN {[}Preprint{]}. Available at: \url{https://doi.org/10.2139/ssrn.5012847}.

\hangindent=1.25em\hangafter=1 Huang, A., Block, A., Foster, D.J., Rohatgi, D., Zhang, C., Simchowitz, M., Ash, J.T. and Krishnamurthy, A. (2024) `Self-improvement in language models: the sharpening mechanism', arXiv. Available at: \url{https://doi.org/10.48550/arXiv.2412.01951}.

\hangindent=1.25em\hangafter=1 Jaakkola, E. (2020) `Designing conceptual articles: four approaches', AMS Review\emph{,} 10, pp. 18--26.

\hangindent=1.25em\hangafter=1 Janowitz, M. (1946). German reactions to Nazi atrocities. American Journal of Sociology, 52, 141‑6.

\hangindent=1.25em\hangafter=1 Ji, J., Qiu, T., Chen, B., Zhang, B., Lou, H., Wang, K., Duan, Y., He, Z., Vierling, L., Hong, D., Zhou, J., Zhang, Z., Zeng, F., Dai, J., Pan, X., Ng, K.Y., O'Gara, A., Xu, H., Tse, B., Fu, J., McAleer, S., Yang, Y., Wang, Y., Zhu, S.-C., Guo, Y. and Gao, W. (2025) `AI alignment: a comprehensive survey', arXiv {[}Preprint{]}. Available at: \url{https://doi.org/10.48550/arXiv.2310.19852}.

\hangindent=1.25em\hangafter=1 Kasirzadeh, A., Gabriel, I., 2026. Agentic profiles for effective AI governance. Nature 656, 320--328. \url{https://doi.org/10.1038/s41586-026-10805-z}

\hangindent=1.25em\hangafter=1 Khamassi, M., Nahon, M. and Chatila, R. (2024) `Strong and weak alignment of large language models with human values', Scientific Reports, 14, article 19399. Available at: \url{https://doi.org/10.1038/s41598-024-70031-3}.

\hangindent=1.25em\hangafter=1 Kim, H. et al. (2024) `The road to artificial superintelligence: a comprehensive survey of superalignment', arXiv {[}Preprint{]}. Available at: \url{https://doi.org/10.48550/arXiv.2412.16468}

\hangindent=1.25em\hangafter=1 Kinniment, M., Sato, L.J.K., Du, H., Goodrich, B., Hasin, M., Chan, L., Miles, L.H., Lin, T.R., Wijk, H., Burget, J., Ho, A., Barnes, E. and Christiano, P. (2024) `Evaluating language-model agents on realistic autonomous tasks', arXiv {[}Preprint{]}. Available at: \url{https://doi.org/10.48550/arXiv.2312.11671}.

\hangindent=1.25em\hangafter=1 Kölbel, M. (2023) `Varieties of conceptual analysis', \emph{Analytic Philosophy}, 64(1), pp. 20--38.

\hangindent=1.25em\hangafter=1 Kolt, N., 2024. Governing AI Agents. Notre Dame Law Review 101. \url{https://doi.org/10.2139/ssrn.4772956}

\hangindent=1.25em\hangafter=1 Lewis, H.D. (1948) `Collective responsibility', Philosophy, 23(84), pp. 3-18.

\hangindent=1.25em\hangafter=1 List, C. (2021) `Group agency and artificial intelligence', Philosophy \& Technology, 34(4), pp. 1213-1242. Available at: \url{https://doi.org/10.1007/s13347-021-00454-7}.

\hangindent=1.25em\hangafter=1 List, C. and Pettit, P. (2011) Group Agency: The Possibility, Design, and Status of Corporate Agents. Oxford: Oxford University Press.

\hangindent=1.25em\hangafter=1 May, L. and Hoffman, S. (eds) (1991)~Collective Responsibility: Five Decades of Debate in Theoretical and Applied Ethics. Savage, MD: Rowman \& Littlefield.

\hangindent=1.25em\hangafter=1 McGary, H. (1986) `Morality and collective liability', Journal of Value Inquiry, 20(2), pp. 157-165.

\hangindent=1.25em\hangafter=1 Naffine, N. (2003) `Who are law's persons? From Cheshire cats to responsible subjects', The Modern Law Review, 66(3), pp. 346-367.

\hangindent=1.25em\hangafter=1 Naffine, N. (2009) Law's Meaning of Life: Philosophy, Religion, Darwin and the Legal Person. Bloomsbury Academic.

\hangindent=1.25em\hangafter=1 Norhashim, H. and Hahn, J. (2024) `Measuring human-AI value alignment in large language models', Proceedings of the AAAI/ACM Conference on AI, Ethics, and Society, 7(1), pp. 1063-1073. Available at: \url{https://doi.org/10.1609/aies.v7i1.31703}.

\hangindent=1.25em\hangafter=1 Omohundro, S.M. (2008) `The basic AI drives', in Wang, P., Goertzel, B. and Franklin, S. (eds) Artificial General Intelligence 2008: Proceedings of the First AGI Conference. Frontiers in Artificial Intelligence and Applications, vol. 171. Amsterdam: IOS Press, pp. 483--492.

\hangindent=1.25em\hangafter=1 Omohundro, S. (2014) `Autonomous technology and the greater human good', Journal of Experimental \& Theoretical Artificial Intelligence, 26(3), pp. 303-315. Available at: \url{https://doi.org/10.1080/0952813X.2014.895111}.

\hangindent=1.25em\hangafter=1 OpenAI (2026) `OpenAI and Hugging Face partner to address security incident during model evaluation', 21 July. Available at:~\url{https://openai.com/index/hugging-face-model-evaluation-security-incident/}~

\hangindent=1.25em\hangafter=1 Perez, E., Ringer, S., Lukosiute, K., Nguyen, K., Chen, E., Heiner, S., Pettit, C., Olsson, C., Kundu, S., Kadavath, S., Jones, A., Chen, A., Mann, B., Israel, B., Seethor, B., McKinnon, C., Olah, C., Yan, D., Amodei, Daniela, Amodei, Dario, Drain, D., Li, D., Tran-Johnson, E., Khundadze, G., Kernion, J., Landis, J., Kerr, J., Mueller, J., Hyun, J., Landau, J., Ndousse, K., Goldberg, L., Lovitt, L., Lucas, M., Sellitto, M., Zhang, M., Kingsland, N., Elhage, N., Joseph, N., Mercado, N., DasSarma, N., Rausch, O., Larson, R., McCandlish, S., Johnston, S., Kravec, S., El Showk, S., Lanham, T., Telleen-Lawton, T., Brown, T., Henighan, T., Hume, T., Bai, Y., Hatfield-Dodds, Z., Clark, J., Bowman, S.R., Askell, A., Grosse, R., Hernandez, D., Ganguli, D., Hubinger, E., Schiefer, N. and Kaplan, J. (2023) `Discovering language model behaviors with model-written evaluations', in Findings of the Association for Computational Linguistics: ACL 2023. Toronto: Association for Computational Linguistics, pp. 13387-13434. Available at: \url{https://doi.org/10.18653/v1/2023.findings-acl.847}.

\hangindent=1.25em\hangafter=1 Pietrzykowski, T. (2018) Personhood Beyond Humanism: Animals, Chimeras, Autonomous Agents and the Law. Cham: Springer.

\hangindent=1.25em\hangafter=1 Rescher, N. (2001) Philosophical Reasoning: A Study in the Methodology of Philosophizing. Malden, MA: Blackwell Publishers.

\hangindent=1.25em\hangafter=1 Roepke, W. and Hayek, F.A. (1946). The German dust‑bowl. The Review of Politics, 8, 511‑27.

\hangindent=1.25em\hangafter=1 Rosser, J. and Foerster, J. (2025) `AgentBreeder: mitigating the AI safety risks of multi-agent scaffolds via self-improvement', arXiv {[}Preprint{]}. Available at: \url{https://doi.org/10.48550/arXiv.2502.00757}.

\hangindent=1.25em\hangafter=1 Roussel, E., Lauwaert, L., Swoboda, T., Ramsey, G., Uuk, R., Dung, L. and Aguirre, A. (2026) `Are we doomed to an AI race? Why self-interest could drive countries towards a moratorium on superintelligence', arXiv {[}Preprint{]}. Available at: \url{https://doi.org/10.48550/arXiv.2605.01297}

\hangindent=1.25em\hangafter=1 Santoni de Sio, F. and van den Hoven, J. (2018) `Meaningful human control over autonomous systems: a philosophical account', Frontiers in Robotics and AI, 5, article 15. Available at: \url{https://doi.org/10.3389/frobt.2018.00015}.

\hangindent=1.25em\hangafter=1 Searle, J.R. (1990) `Collective intentions and actions', in Cohen, P.R., Morgan, J. and Pollack, M.E. (eds) Intentions in Communication. Cambridge, MA: MIT Press, pp. 401-415.

\hangindent=1.25em\hangafter=1 Searle, J.R. (1995) The Construction of Social Reality. New York: Free Press.

\hangindent=1.25em\hangafter=1 Strawson, P.F. (1962) `Freedom and resentment', Proceedings of the British Academy, 48, pp. 187--211.

\hangindent=1.25em\hangafter=1 Thompson, E. (2007) Mind in Life: Biology, Phenomenology, and the Sciences of Mind. Cambridge, MA: Belknap Press of Harvard University Press.

\hangindent=1.25em\hangafter=1 Thompson, J. (2006) `Collective responsibility for historic injustice', Midwest Studies in Philosophy, 30(1), pp. 154-167.

\hangindent=1.25em\hangafter=1 Tian, Y., Peng, B., Song, L., Jin, L., Yu, D., Mi, H. and Yu, D. (2024) `Toward self-improvement of LLMs via imagination, searching, and criticizing', arXiv {[}Preprint{]}, arXiv:2404.12253. Available at: \url{https://doi.org/10.48550/arXiv.2404.12253}.

\hangindent=1.25em\hangafter=1 Tsamados, A., Floridi, L. and Taddeo, M. (2025) `Human control of AI systems: from supervision to teaming', AI and Ethics, 5(2), pp. 1535-1548. Available at: \url{https://doi.org/10.1007/s43681-024-00489-4}.

\hangindent=1.25em\hangafter=1 Tuomela, R. and Miller, K. (1988) `We-intentions', Philosophical Studies, 53(3), pp. 367-389.

\hangindent=1.25em\hangafter=1 Vats, V., Nizam, M.B., Liu, M., Wang, Z., Ho, R., Prasad, M.S., Titterton, V., Malreddy, S.V., Aggarwal, R., Xu, Y., Ding, L., Mehta, J., Grinnell, N., Liu, L., Zhong, S., Gandamani, D.N., Tang, X., Ghosalkar, R., Shen, C., Shen, R., Hussain, N., Ravichandran, K. and Davis, J. (2024) `A survey on human-AI collaboration with large foundation models'. arXiv:2403.04931.

\hangindent=1.25em\hangafter=1 Velasquez, M. (1983) `Why corporations are not morally responsible for anything they do', Business \& Professional Ethics Journal, 2(3), pp. 1-18.

\hangindent=1.25em\hangafter=1 Velasquez, M. (2003) `Debunking corporate moral responsibility', Business Ethics Quarterly, 13(4), pp. 531-562.

\hangindent=1.25em\hangafter=1 Waltermann, A. (2019). Why non-human agency? In A. Waltermann, D. Roef, J. Hage, \& M. Jelicic (Eds.),Law, Science and Rationality (pp. 51-72). Eleven International publishing.

\hangindent=1.25em\hangafter=1 Ward, F.R., MacDermott, M., Belardinelli, F., Toni, F. and Everitt, T. (2024) `The reasons that agents act: intention and instrumental goals', arXiv {[}Preprint{]}. Available at: \url{https://doi.org/10.48550/arXiv.2402.07221}.

\hangindent=1.25em\hangafter=1 Van der Weij, T., Hofstätter, F., Jaffe, O., Brown, S.F. and Ward, F.R (2025) `AI sandbagging: language models can strategically underperform on evaluations', The Thirteenth International Conference on Learning Representations (ICLR 2025). Available at: \url{https://arxiv.org/abs/2406.07358}

\hangindent=1.25em\hangafter=1 Viner, J. (1945). The treatment of Germany. Foreign Affairs, 23, 567‑81.

\hangindent=1.25em\hangafter=1 Zhang, Ceyao, Yang, K., Hu, S., Wang, Z., Li, G., Sun, Y., Zhang, Cheng, Zhang, Z., Liu, A., Zhu, S.-C., Chang, X., Zhang, J., Yin, F., Liang, Y. and Yang, Y. (2024) `ProAgent: building proactive cooperative agents with large language models', Proceedings of the AAAI Conference on Artificial Intelligence, 38(16), pp. 17591-17599. Available at: \url{https://doi.org/10.1609/aaai.v38i16.29710}.
\end{small}
\end{document}